\documentclass[letterpaper]{article} 
\usepackage{aaai2027}  
\usepackage[hyphens]{url}  
\usepackage{graphicx} 
\usepackage{natbib}  
\usepackage{caption} 
\usepackage{algorithm}
\usepackage{algorithmic}

\usepackage{newfloat}
\usepackage{listings}
\DeclareCaptionStyle{ruled}{labelfont=normalfont,labelsep=colon,strut=off} 
\floatstyle{ruled}
\newfloat{listing}{tb}{lst}{}
\floatname{listing}{Listing}

\usepackage{booktabs}

\usepackage{xcolor}
\usepackage{xspace}
\definecolor{bannerBlue}{HTML}{547bb4}
\definecolor{bannerGreen}{HTML}{84ba42}
\definecolor{bannerRed}{HTML}{c0321a}

\definecolor{motivationGreen}{HTML}{4ea72e}
\definecolor{motivationRed}{HTML}{c00000}
\usepackage{mdframed}
\usepackage{bbding}
\usepackage{pifont}
\usepackage[export]{adjustbox}
\usepackage{threeparttable}
\usepackage{subcaption}
\usepackage{multirow}
\usepackage{enumitem}
\usepackage{colortbl}
\usepackage{arydshln}
\usepackage{amsfonts}
\usepackage{amssymb}
\usepackage{bm}
\usepackage{mathrsfs}  
\usepackage{makecell}
\usepackage{amsmath}
\usepackage{tabularx}
\usepackage{array}           

\newcommand{\cref}[1]{\ref{#1}}
\newcommand{\Cref}[1]{\ref{#1}}

\definecolor{mygray}{gray}{.9}
\definecolor{mygreen}{RGB}{93,173,85}
\definecolor{mywarning}{RGB}{233,144,61}

\definecolor{DarkBlue}{RGB}{64,101,149}
\definecolor{azure}{rgb}{0.0, 0.5, 1.0}
\definecolor{gray}{rgb}{0.3, 0.3, 0.3}
\definecolor{DarkGreen}{RGB}{42,110,63}
\definecolor{DarkYellow}{RGB}{191,144,0}
\definecolor{DarkRed}{rgb}{0.6, 0, 0} 

\newcolumntype{Y}{>{\centering\arraybackslash}X}

\newcolumntype{x}[1]{>{\centering\arraybackslash}p{#1pt}}
\newcolumntype{I}{!{\vrule width 1pt}}

\makeatletter
\newcommand{\thickhline}{%
    \noalign {\ifnum 0=`}\fi \hrule height 1pt
    \futurelet \reserved@a \@xhline
}

\newcommand{\ourmethod}{{\fontfamily{lmtt}\selectfont \textbf{RippleNet}}\xspace}

\title{Structured Local Differential Modeling for AI-Generated Image Detection}

\author{
    Jiazhen Yang\textsuperscript{\rm 1},
    Ruijin Jin\textsuperscript{\rm 2},
    Junjun Zheng\textsuperscript{\rm 2},
    Xiangheng Kong\textsuperscript{\rm 2},
    Zunlei Feng\textsuperscript{\rm 1},
    Jie Lei\textsuperscript{\rm 3},
}
\affiliations{
    \textsuperscript{\rm 1} Zhejiang University
    \textsuperscript{\rm 2} Taobao \& Tmall Group of Alibaba 
    \textsuperscript{\rm 3} Zhejiang University of Technology
}

\begin{document}

\maketitle

\begin{abstract}
    The rapid advancement of AI-generated content has made the reliable detection of generated images an increasingly critical challenge. Existing detection methods are often dominated during training by semantically salient components with high signal-to-noise ratios (SNRs), thereby suppressing subtler forensic cues associated with the underlying generation mechanisms and embedded in low-level statistical structures. From an information-theoretic perspective, we present a key insight: effective detection in the low-level statistical space requires mitigating the dominance of semantic components while emphasizing and amplifying responses to low-SNR forgery traces.
    Building on this insight, we propose \ourmethod{}, an AI-generated image detection framework based on local differential signals. \ourmethod{} adaptively identifies forgery-sensitive regions and constructs multi-directional, multi-scale differential representations within local neighborhoods, explicitly characterizing anomalous patterns in neighborhood statistics. More importantly, we refine the attention mechanism to operate within the local differential representation space, enabling the model to establish explicit dependencies at a finer statistical granularity. This design facilitates the capture of pixel-level forgery traces that are difficult to model using conventional convolutions or image-wide patch-level attention. Extensive experiments on multiple public benchmarks and under cross-generator evaluation settings demonstrate that \ourmethod{} achieves consistently competitive performance.
\end{abstract}

\section{Introduction}
\label{sec:introduction}

Recent advances in image generation~\cite{cao2025controllable} have substantially improved the photorealism of synthetic images, while heightening concerns over content authenticity and making generalizable generated image detection increasingly important~\cite{xu2025recent, mahara2026methods}.

\begin{figure}[t]
    \centering
    \includegraphics[width=0.48\textwidth]{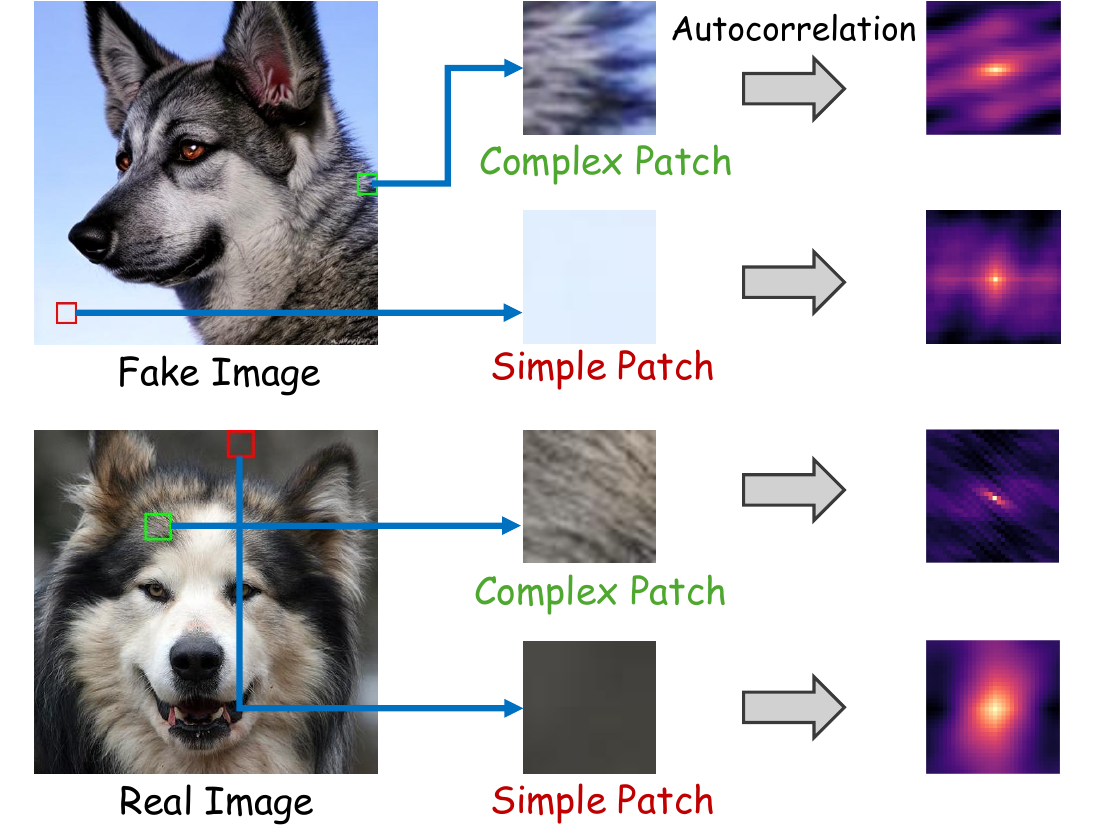}
    \vspace{-10pt}
    \caption{\textbf{Autocorrelation differences between real and generated images in \textcolor{motivationGreen}{complex} and \textcolor{motivationRed}{simple} patches.} Based on the Wiener--Khinchin theorem, real images exhibit clear directional structures in complex regions and slight noise fluctuations in smooth regions, whereas generated images often lack directional continuity and display overly smoothed, radially symmetric responses in smooth regions.
    }
    \label{fig:motivation}
    \vspace{-14pt}
\end{figure}

\begin{figure*}[t]
\centering
\includegraphics[width=0.88\linewidth]{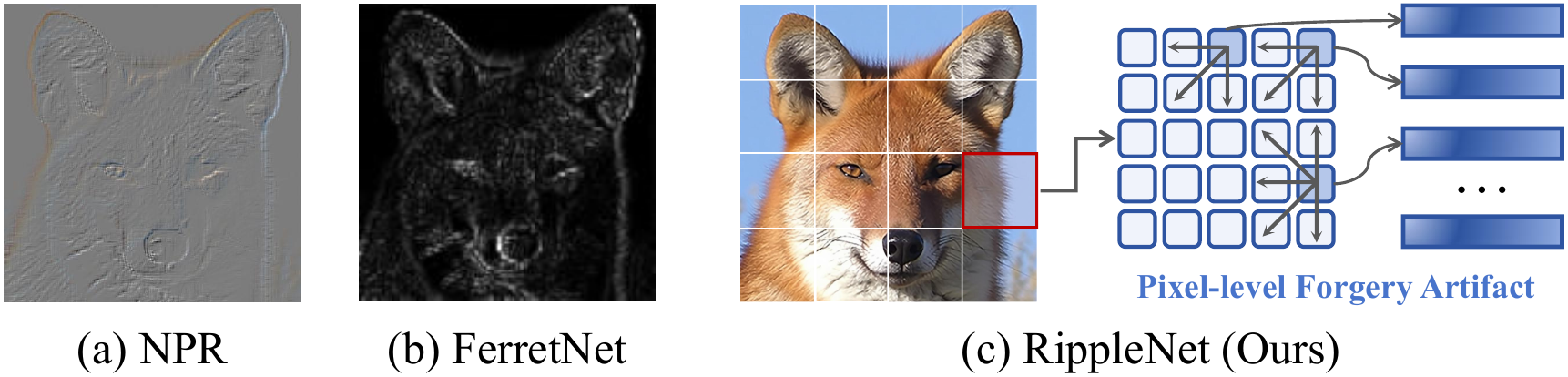}
\vspace{-5pt}
\caption{\textbf{Comparison of forgery representations from NPR~\cite{tan2024rethinking}, FerretNet~\cite{liang2025ferretnet}, and \ourmethod{}.} NPR and FerretNet construct image-level cues from local correlations or filtered noise, which may retain semantic textures and weaken subtle artifacts during convolutional aggregation. In contrast, \ourmethod{} encodes neighborhood differential responses as independent tokens and models their dependencies directly in the token space.}
\label{fig:banner}
\vspace{-15pt}
\end{figure*}

Existing methods generally operate in either the semantic space or the low-level statistical space. Semantic methods~\cite{ojha2023towards,zhang2025towards,zhou2026forensicconcept} exploit high-level representations learned by deep networks or large-scale pretrained models~\cite{radford2021learning} to distinguish generated images from real ones. Although effective under in-distribution settings, their discriminative cues may become entangled with training semantics, limiting transfer to unseen generators and increasingly realistic content. In contrast, low-level statistical methods~\cite{tan2024rethinking,liang2025ferretnet,zhong2023patchcraft,qi2026difference} target generation-induced anomalies in spectral distributions, noise patterns, and local textures. As these cues are less dependent on image semantics and more closely related to synthesis mechanisms, they are generally considered more promising for cross-generator generalization.

However, statistical space detection still faces a key challenge: forgery artifacts are often weak, local, and low-SNR, making them susceptible to dominant semantic structures during whole-image aggregation and end-to-end optimization. As illustrated in \cref{fig:banner}, NPR~\cite{tan2024rethinking} models local relationships, while FerretNet~\cite{liang2025ferretnet} extracts noise responses. Although both methods capture useful forensic cues, their image-level representations still retain prominent content-related textures. From an information-theoretic perspective~\cite{tishby2000information}, when high-SNR semantic structures coexist with low-SNR forgery artifacts, optimization tends to favor the former because they are easier to fit and reduce the training loss more rapidly. This bias may suppress the modeling of weak forgery artifacts and encourage content-dependent shortcuts, thereby limiting cross-generator generalization. Based on this, we formulate the central hypothesis of this work:
The key to low-level statistical space generalization detection lies in mitigating the dominance of high-SNR semantic components while enhancing the response to low-SNR forgery artifacts, thereby compelling the model to learn more transferable cues associated with generative mechanisms.


Based on this viewpoint, we revisit the differences between real and generated images from the perspective of local structures and differential statistics. As shown in \cref{fig:motivation}, real images, shaped by physical imaging pipelines, tend to exhibit more consistent and reproducible neighborhood statistics. In complex texture regions, their residual responses preserve stable directional correlations. In simple regions, sensor noise still produces weak yet repeatable fluctuations despite the absence of salient textures. In contrast, although generative models can faithfully reproduce global semantics and prominent textures, they often struggle to maintain statistical consistency across multiple directions and scales during local reconstruction. This limitation gives rise to characteristic anomalies, such as weakened neighborhood correlations and excessive smoothing, which can be quantified through local differential correlation statistics.

Building on this insight, we propose \ourmethod{}, a generated image detection framework centered on low-SNR differential signals. As shown in \cref{fig:banner}, \ourmethod{} first localizes regions sensitive to forgery artifacts and then models the propagation and continuity of neighborhood differences across multiple directions and scales, jointly capturing local structural relations and statistical dependencies. To further preserve weak forgery responses during global aggregation, we encode local differential artifacts as independent tokens and perform attention-based modeling directly in this representation space. This allows the model to establish cross-location dependencies at a finer statistical granularity, reduce interference from semantic components, and improve generalization to unseen generative architectures. We make the following key contributions:

\begin{itemize}[leftmargin=*]

\item[\ding{182}] \textbf{New perspective on statistical space forgery detection.} From the perspectives of generative mechanisms and information-theoretic bias, we argue that suppressing semantic dominance and emphasizing low-SNR statistical discrepancies can improve cross-generator generalization.

\item[\ding{183}] \textbf{Novel forgery artifact modeling framework.} We propose \ourmethod{}, a forgery detection method tailored to low-level differential statistics, enhancing forgery artifact representations by modeling neighborhood residual relationships and cross location interactions through attention.

\item[\ding{184}] \textbf{Experimental validation of generalizable detection.} We evaluate our method on multiple public cross generator benchmarks and show that it exhibits more stable generalization under different generative paradigms.
\end{itemize}

\section{Related Work}

\noindent{\textbf{Synthetic Detection in Semantic Space.}}
Semantic methods leverage high-level representations from deep networks or pretrained vision--language models such as CLIP~\cite{radford2021learning}. UnivFD~\cite{ojha2023towards} performs detection using frozen CLIP features, while C2P-CLIP~\cite{tan2025c2p} improves feature--semantic alignment through class prompts. Effort~\cite{yan2024effort} separates semantic and forgery-related information into orthogonal subspaces, and VIBNet~\cite{zhang2025towards} employs a variational information bottleneck to suppress irrelevant information. Although effective on multiple benchmarks, these methods may couple discriminative cues with training semantics, limiting generalization to unseen generators.

\noindent{\textbf{Synthetic Detection in Low-Level Statistical Space.}}
Low-level methods detect generation-induced anomalies in geometry and color~\cite{sarkar2024shadows,jia2025secret}, spectral statistics~\cite{tan2024frequency,karageorgiou2025any}, local correlations~\cite{tan2024rethinking,li2025improving,yuan2026mlep}, and noise or texture patterns~\cite{zhong2023patchcraft,chen2024single,liang2025ferretnet}. Reconstruction-based methods such as DIRE~\cite{wang2023dire} and STD-FD~\cite{lou2025std} characterize distributional deviations, while Difference-in-Difference~\cite{qi2026difference} further exploits second-order reconstruction-error differences. Although less dependent on semantics, these cues may still retain content-related structures, limiting cross-domain generalization.

\noindent{\textbf{Semantic--Artifact Feature Integration.}}
Recent studies combine semantic representations with low-level forensic cues. AIDE~\cite{yan2024sanity} integrates CLIP features with noise patterns from low- and high-frequency patches, while CO-SPY~\cite{cheng2025co} combines semantic and reconstruction-error features. More recently, ForensicConcept~\cite{zhou2026forensicconcept} organizes decision-critical regions into transferable forensic concepts and injects them across backbones, whereas TranX-Adapter~\cite{wang2026tranx} improves artifact--semantic interaction in MLLMs through optimal-transport and cross-attention modules. These methods demonstrate the complementarity of semantic and forensic cues, although their effectiveness depends on aligning and fusing heterogeneous representations.

\section{Motivation}

\subsection{Frequency Dependent Reconstruction Biases}

Although modern generative models can reproduce global semantics and coarse structures with high fidelity, their synthesis processes may still introduce frequency dependent biases in fine-scale components. For convolutional generators, transposed convolution may produce periodic spectral replicas because of uneven overlap, whereas interpolation followed by convolution tends to suppress fine-scale responses through its inherent low-pass behavior. Despite different signatures, both mechanisms can alter the frequency organization and local structural consistency of synthesized images.

Diffusion models exhibit a related imbalance during noising and denoising. Since fine scale components generally carry less energy, they enter a low-SNR regime earlier and must be reconstructed from less reliable observations. Let $\theta$ denote the model parameters, $\mathcal{L}$ the generation objective, and $\|\nabla_\theta \mathcal{L}\|_f$ the effective gradient contribution associated with frequency component $f$. Under a simplified frequency-wise view, its magnitude can be approximately characterized as

\begin{equation}
\|\nabla_\theta \mathcal{L}\|_f
\propto
\mathrm{SNR}_f^{1/2}.
\end{equation}

This relation indicates that low-SNR frequency components provide weaker effective optimization signals and are therefore more likely to be underrepresented during learning. Consequently, generated images may preserve global content while exhibiting reduced consistency in fine-scale structures.

Although these biases originate from different generative mechanisms, they may share a common manifestation in abnormal directional and scale-dependent local variations. 

\subsection{An Information-Theoretic View of Representation Imbalance}

The preceding analysis explains how generative processes may leave weak inconsistencies in fine-scale structures. However, these cues may not be effectively preserved during detector training because they compete with more salient and easily fitted content structures.
From the information-bottleneck~(IB) perspective, representation learning seeks $Z=h_{\theta}(X)$ that remains predictive of the label $Y$ while discarding task-irrelevant input information:

\begin{equation}
\max_{\theta}\ I(Z;Y)-\beta I(Z;X),
\end{equation}
where $\beta$ controls the compression--prediction trade-off. Although this objective does not explicitly favor any particular input component, finite-data optimization tends to prioritize factors that are strongly correlated with the training labels and easier to fit. In generated-image detection, these factors often correspond to salient content and coarse structural regularities, whereas generation-related discrepancies are typically weaker and less prominent.
Let $X_s$ and $X_a$ denote the dominant structural and weak artifact-related components, respectively. A shortcut-dominated representation can be qualitatively characterized by
\begin{equation}
\left\|\nabla_{X_s}h_{\theta}(X)\right\|
\gg
\left\|\nabla_{X_a}h_{\theta}(X)\right\|,
\end{equation}
indicating substantially greater sensitivity to structural variations than to artifact-related ones. Such imbalance may cause the detector to encode content correlations specific to the training distribution, limiting transfer to unseen generators.

Taken together, generative models may leave subtle statistical inconsistencies that are often overshadowed by dominant content cues during detector training. This motivates a representation strategy that emphasizes transferable forgery evidence while reducing reliance on global image content. Further theoretical analysis is provided in \textbf{Appendix}.

\begin{figure*}
\centering
\includegraphics[width=\linewidth]{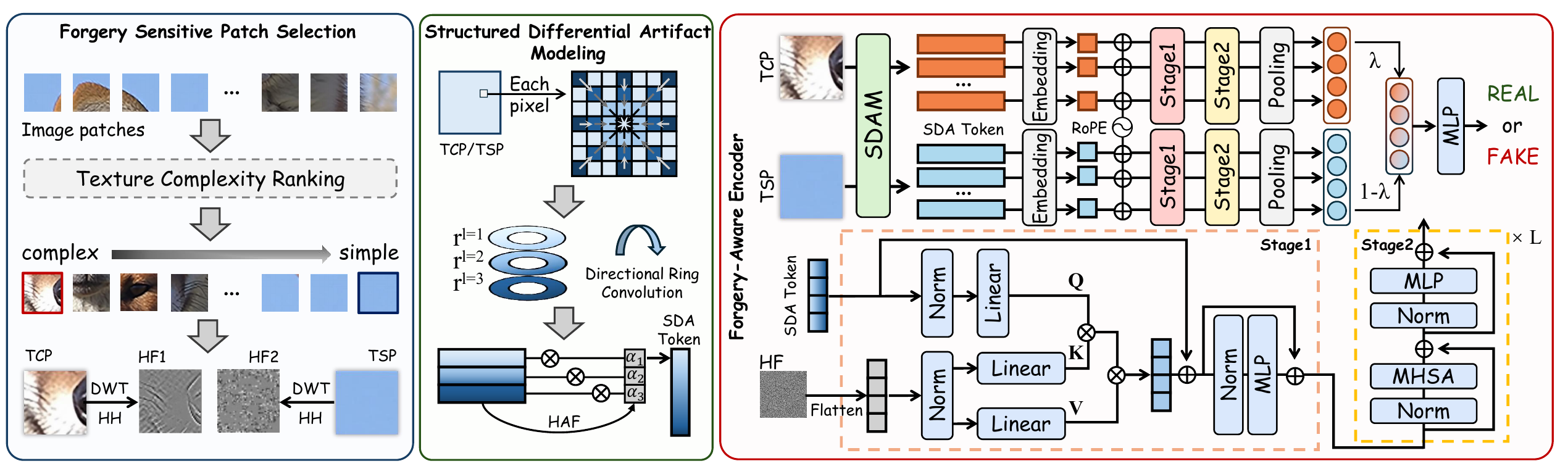}
\vspace{-15pt}
\captionsetup{font=small}
\caption{\textbf{Overview of \ourmethod{}.} \textcolor{bannerBlue}{\textbf{Forgery Sensitive Patch Selection}} module analyzes local texture complexity to select Texture-Complex Patches~(TCPs) and Texture-Simple Patches~(TSPs) that provide complementary forgery evidence, and extracts their high-frequency components via DWT for subsequent modeling.
\textcolor{bannerGreen}{\textbf{Structured Differential Artifact Modeling}} module constructs multi-directional and multi-scale differential representations at each spatial location and refines them through Directional Ring Convolution~(DRC) and Hierarchical Attention Fusion~(HAF).
\textcolor{bannerRed}{\textbf{Forgery Aware Encoder}} independently encodes the two patch types, incorporates high-frequency guidance as a frequency-domain prior, and employs multi-head self-attention to progressively model dependencies among differential tokens.}
\vspace{-10pt}
\label{fig:framework}
\end{figure*}

\section{Method}

To capture subtle generation related traces that may be obscured by dominant content structures, we propose \ourmethod{}, a detection framework built on local differential statistics. \ourmethod{} first selects complementary regions with different texture complexities, then constructs multi-directional and multi-scale neighborhood differential representations and incorporates frequency priors to model their dependencies. By consolidating local anomalous responses into coherent discriminative evidence, \ourmethod{} reduces direct reliance on global content structures. The overall framework is illustrated in \cref{fig:framework}.

\subsection{Forgery Sensitive Patch Selection}

As observed in \cref{fig:motivation}, real and fake images exhibit distinct local differential statistics across regions with different texture complexity. Whole-image modeling may mix weak artifact cues with dominant structural responses during aggregation~\cite{tan2024rethinking, liang2025ferretnet}. We therefore introduce Forgery Sensitive Patch Selection~(FSPS), which selects texture-complex and texture-simple patches as complementary inputs for subsequent differential modeling.

Given an input image $I$, we partition it into a set of non-overlapping patches $\mathcal{P}$. For each patch $p \in \mathcal{P}$, we measure its texture complexity using the total variation~(TV) energy along four directions $(\rightarrow,\downarrow,\searrow,\nearrow)$:

\begin{flalign}
\label{eq:tv_left}
& TV(p) = \sum_{(i,j)\in p} \Big( |I_{i,j} - I_{i,j+1}| + |I_{i,j} - I_{i+1,j}| \notag \\
& \quad + |I_{i,j} - I_{i+1,j+1}| + |I_{i,j} - I_{i+1,j-1}| \Big), 
\end{flalign}

Here, $I_{i,j}$ denotes the grayscale intensity at spatial location $(i,j)$. We rank all patches by $\mathrm{TV}(p)$ and select $m$ patches from each end of the ranking. The $m$ highest-scoring patches form the texture-complex patch~(TCP) set $\mathcal{P}_{\mathrm{TC}}$, while the $m$ lowest-scoring patches form the texture-simple patch~(TSP) set $\mathcal{P}_{\mathrm{TS}}$. These complementary texture regimes guide subsequent modeling toward regions where forgery cues may be more evident, providing a more discriminative prior for differential representation learning.

\subsection{Structured Differential Artifact Modeling}

To enhance the detector's sensitivity to forgery cues, we introduce Structured Differential Artifact Modeling~(SDAM), which transforms local variations into structured representations by jointly modeling directional organization and scale-dependent responses. This design highlights deviations in local geometric consistency and energy distribution.

\noindent{\textbf{Multi-directional and Multi-scale Residual Construction.}}
Given an input patch $p \in \mathbb{R}^{h \times w}$, SDAM constructs a radially expanding differential sequence at each spatial location $x_{i,j}$ along eight directions. We define
$\mathcal{D}
=
\{\uparrow,\downarrow,\rightarrow,\leftarrow,
\nearrow,\nwarrow,\searrow,\swarrow\}$,
where each direction $d_k \in \mathcal{D}$ corresponds to a unit offset vector
$(\Delta_i^{(k)},\Delta_j^{(k)})$, with $k \in \{0,\ldots,7\}$. Given a maximum radial step $L$, the residual at step $l \in \{1,\ldots,L\}$ along direction $d_k$ is defined as
\begin{equation}
r_{i,j}^{(k,l)}
=
I_{i+\Delta_i^{(k)}l,\,j+\Delta_j^{(k)}l}
-
I_{i+\Delta_i^{(k)}(l-1),\,j+\Delta_j^{(k)}(l-1)}.
\end{equation}

This operation measures incremental variations between adjacent radial positions, capturing how local dependencies evolve across directions and scales. Concatenating all residuals yields the local differential descriptor
\begin{equation}
\mathbf{r}_{i,j}
=
[r_{i,j}^{(k,l)}]_{k=0,\ldots,7;\,l=1,\ldots,L}
\in \mathbb{R}^{8\times L}.
\end{equation}

\noindent{\textbf{Directional Ring Convolution.}}
Single-direction differences are insufficient to capture disruptions in directional consistency and local symmetry. We therefore introduce Directional Ring Convolution~(DRC) to model cyclic dependencies among the eight directions. At scale $l$, it is defined as

\begin{equation}
\tilde{r}_{i,j}^{(k,l)}
=
\sum_{t=-\lfloor \mathcal{K}/2 \rfloor}^{\lfloor \mathcal{K}/2 \rfloor}
w_t\,
r_{i,j}^{((k+t)\bmod 8,\,l)},
\end{equation}
where $\mathcal{K}$ is the kernel size and $w_t$ denotes learnable weights. The modulo operation preserves circular adjacency in the directional domain, producing direction-enhanced differential features
$
\tilde{\mathbf{r}}_{i,j}
=
[\tilde{r}_{i,j}^{(k,l)}]_{k=0,\ldots,7;\,l=1,\ldots,L}$.

\noindent{\textbf{Hierarchical Attention Fusion.}}
Different neighborhood scales provide complementary artifact evidence. Short-range residuals are sensitive to subtle texture variations, whereas larger neighborhoods better reflect deviations in structural consistency. We therefore employ Hierarchical Attention Fusion~(HAF) to adaptively aggregate cross-scale information. Let $\tilde{\mathbf{r}}_{i,j}^{(l)} \in \mathbb{R}^{C}$ denote the feature at scale $l$. We stack the features across all scales into $\mathbf{A}$ and generate scale-attention weights using a lightweight MLP:
\begin{equation}
\boldsymbol{\alpha}
=
\mathrm{Softmax}\!\Big(
W_2\,\phi(W_1\mathbf{A}+b_1)+b_2
\Big),
\end{equation}
where $\phi(\cdot)$ denotes GELU activation, $W_1$ and $W_2$ are learnable parameters, $b_1$ and $b_2$ are bias terms. The fused pixel-level forgery embedding is given by
\begin{equation}
\hat{\mathbf{r}}_{i,j}
=
\sum_{l=1}^{L}
\boldsymbol{\alpha}_l
\odot
\tilde{\mathbf{r}}_{i,j}^{(l)}
\in \mathbb{R}^{D},
\end{equation}
where $\odot$ denotes channel-wise weighting. HAF adaptively emphasizes the most informative scales according to local content, yielding a more discriminative representation.

\begin{figure*}[t]
  \centering
  \begin{minipage}[t]{0.68\textwidth}
    \captionsetup{font=small}
    \captionof{table}{Comparison of \ourmethod{} and other forgery detection models on the GenImage Benchmark in terms of ACC (\%). All methods are trained on SDv1.4.
    \textbf{Bold} indicates the best result, and \underline{underline} denotes the second-best.}
    \centering
    \scriptsize
    \setlength\tabcolsep{3pt}
    \renewcommand\arraystretch{1.1}
    \begin{adjustbox}{width=\linewidth, valign=t}
      \begin{tabular}{r|c|c|c|c|c|c|c|c|c}
      \thickhline
      \rowcolor{gray!20}
      Methods
      & Midj & SDv1.4 & SDv1.5 & ADM & GLIDE
      & Wukong & VQDM & BigGAN & Avg. \\
      \hline

      F3Net~\cite{qian2020thinking}
      & 77.9 & 99.0 & 99.1 & 51.2 & 54.9 & 97.9 & 59.0 & 49.2 & 73.5 \\

      \rowcolor{gray!10}
      FreqNet~\cite{tan2024frequency}
      & 89.6 & 98.8 & 98.6 & 66.8 & 96.5 & 97.3 & 75.8 & 81.4 & 88.1 \\

      FatFormer~\cite{liu2024forgery}
      & 92.7 & 100.0 & 99.9 & 75.9 & 88.0 & 99.9 & 98.8 & 55.8 & 88.9 \\

      \rowcolor{gray!10}
      NPR~\cite{tan2024rethinking}
      & 81.0 & 98.2 & 97.9 & 76.9 & 89.8 & 96.9 & 84.1 & 84.2 & 88.6 \\

      DRCT~\cite{chen2024drct}
      & 91.5 & 95.0 & 94.4 & 79.4 & 89.2 & 94.7 & 90.0 & 81.7 & 89.5 \\

      \rowcolor{gray!10}
      AIDE~\cite{yan2024sanity}
      & 79.4 & 99.7 & 99.8 & 78.5 & 91.8 & 98.7 & 80.3 & 66.9 & 86.9 \\

      VIBNet~\cite{zhang2025towards}
      & 88.1 & 99.6 & 99.2 & 73.8 & 74.2 & 98.3 & 89.3 & 72.6 & 86.9 \\

      \rowcolor{gray!10}
      Effort~\cite{yan2024effort}
      & 82.4 & 99.8 & 99.8 & 78.7 & 93.3 & 97.4 & 91.7 & 77.6 & 91.1 \\

      FerretNet~\cite{liang2025ferretnet}
      & 88.3 & 98.6 & 98.4 & 74.5 & 97.4 & 97.8 & 81.6 & 80.9 & 89.7 \\
     
      \rowcolor{gray!10}
      TranX-Adapter~\cite{wang2026tranx}
      & 94.6 & 96.4 & 96.4 & 87.0 & 88.0 & 94.9 & 90.1 & 85.9 & 91.9 \\

      CKNNA~\cite{zhou2026forensicconcept}
      & 95.0 & 99.6 & 99.4 & 69.2 & 85.1 & 99.6 & 94.3 & 94.1 & \underline{92.0} \\

      \rowcolor[HTML]{D7F6FF}
      \ourmethod{}
      & 89.2 & 98.8 & 98.6 & 93.6 & 96.6 & 97.9 & 92.2 & 88.2 & \textbf{94.4} \\
      \thickhline
      \end{tabular}
    \end{adjustbox}
    \label{tab:table1}
  \end{minipage}
  \hfill
  \begin{minipage}[t]{0.3\textwidth}
    \vspace{0pt}
    \centering
    \includegraphics[width=\linewidth]{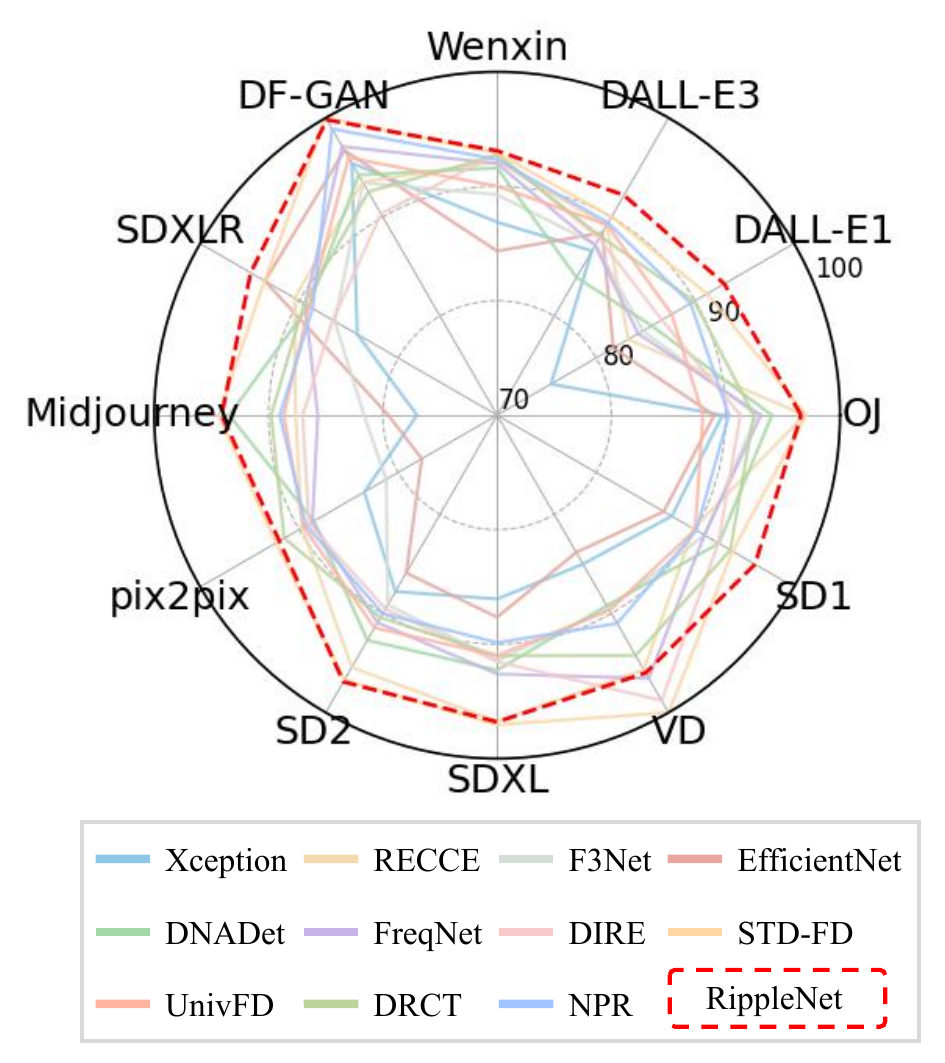}
    \captionsetup{font=small}
    \vspace{-20pt}
    \captionof{figure}{Comparison of \ourmethod{} and other models on DeepfaceGen.}
    \label{fig:radar}
  \end{minipage}
  \vspace{-10pt}
\end{figure*}

\subsection{Forgery Aware Encoder}

To aggregate local differential artifacts into global discriminative evidence, we design a Forgery Aware Encoder~(FAE). Since artifact cues are often reflected by the spatial co-occurrence and consistency of neighborhood differential patterns rather than isolated responses, FAE models dependencies across spatial locations based on representation similarity. Two encoders with identical architectures but separate parameters are used for $\mathcal{P}_{\mathrm{TC}}$ and $\mathcal{P}_{\mathrm{TS}}$, enabling specialized modeling under distinct texture regimes.

\noindent{\textbf{Differential Artifact Tokenization.}}
Given the artifact representation $\hat{\mathbf{r}}_{i,j}$ produced by SDAM, we project and serialize the local descriptors into tokens $\mathbf{E}_s \in \mathbb{R}^{N \times D}$. A 2D rotary positional embedding~(RoPE) is applied to preserve spatial continuity and relative positional relationships.

\noindent{\textbf{Frequency-Guided Cross-Attention.}}
To enhance sensitivity to high-frequency biases induced by generative processes, we introduce Frequency-Guided Cross-Attention~(FGCA) in the first encoder layer. For each patch, DWT is applied to extract the high-frequency subband $HH$, which is resized and flattened into a frequency embedding $\mathbf{E}_f \in \mathbb{R}^{N \times D}$ aligned with $\mathbf{E}_s$. The two embeddings are projected as
\begin{equation}
Q_s=\mathbf{E}_sW_Q, K_f=\mathbf{E}_fW_K, V_f=\mathbf{E}_fW_V,
\end{equation}
where $W_Q,W_K,W_V\in\mathbb{R}^{D\times d_{\mathrm{model}}}$. The cross-attention output is defined as
\begin{equation}
\mathbf{H}
=
\mathrm{Softmax}\!\left(
\frac{Q_sK_f^{\top}}{\sqrt{d_{\mathrm{model}}}}
\right)V_f.
\end{equation}

This mechanism enables differential tokens to attend to high-frequency responses, enhancing sensitivity to spectral anomalies and fine-grained distortions.

\noindent{\textbf{Relational Self-Attention.}}
Following frequency interaction, FAE stacks multi-head self-attention~(MHSA) layers to model dependencies among differential tokens. By establishing interactions across locations according to representation similarity, MHSA captures the co-occurrence and consistency of artifact patterns. This preserves fine-grained spatial relationships during encoding and reduces the attenuation of weak artifact cues during aggregation.

\noindent{\textbf{Dual-Branch Fusion and Classification.}}
The two branches independently encode the texture-complex and texture-simple patch sets, producing
\begin{equation}
\mathcal{H}_{\mathrm{TC}}
=
\{\mathbf{H}_{\mathrm{TC}}^{(1)},\ldots,\mathbf{H}_{\mathrm{TC}}^{(m)}\},
\qquad
\mathcal{H}_{\mathrm{TS}}
=
\{\mathbf{H}_{\mathrm{TS}}^{(1)},\ldots,\mathbf{H}_{\mathrm{TS}}^{(m)}\}.
\end{equation}

We first apply mean pooling over the token sequence of each patch and then average across patches:
\begin{equation}
\mathbf{z}_{\mathrm{TC}}
=
\frac{1}{m}\sum_{q=1}^{m}
\mathrm{Pool}\!\left(\mathbf{H}_{\mathrm{TC}}^{(q)}\right),
\qquad
\mathbf{z}_{\mathrm{TS}}
=
\frac{1}{m}\sum_{q=1}^{m}
\mathrm{Pool}\!\left(\mathbf{H}_{\mathrm{TS}}^{(q)}\right).
\end{equation}
A learnable coefficient $\lambda\in(0,1)$ balances the complementary contributions of the two texture regimes:
\begin{equation}
\mathbf{z}
=
\lambda\mathbf{z}_{\mathrm{TC}}
+
(1-\lambda)\mathbf{z}_{\mathrm{TS}},
\qquad
\lambda=\sigma(\beta),
\end{equation}
where $\beta$ is learnable and $\sigma(\cdot)$ denotes the Sigmoid function. The fused representation is fed into the classification head:
\begin{equation}
\hat{y}=\mathrm{MLP}(\mathbf{z}),
\end{equation}
yielding the final real-versus-fake prediction.

\begin{table*}[t]\small
\centering
\captionsetup{font=small}
\caption{ACC (\%) and AP (\%) comparison of \ourmethod{} and other forgery detection models on the  DiffusionForensics dataset.
All methods are trained on GenImage/SDv1.4.
\textbf{Bold} indicates the best result, and \underline{underline} denotes the second-best.}
\vspace{-5pt}
\label{tab:df_table}
{\scriptsize
\resizebox{\linewidth}{!}{
\setlength\tabcolsep{3pt}
\renewcommand\arraystretch{1.15}
\begin{tabular}{r|cc|cc|cc|cc|cc|cc|cc|cc|cc}
\hline\thickhline
\rowcolor{gray!20}
 & \multicolumn{2}{c|}{ADM}
 & \multicolumn{2}{c|}{DDPM}
 & \multicolumn{2}{c|}{IDDPM}
 & \multicolumn{2}{c|}{LDM}
 & \multicolumn{2}{c|}{PNDM}
 & \multicolumn{2}{c|}{VQ-Diffusion}
 & \multicolumn{2}{c|}{SD1}
 & \multicolumn{2}{c|}{SD2}
 & \multicolumn{2}{c}{Mean} 
\\
\cline{2-19}
\rowcolor{gray!20}
Methods
& ACC & AP & ACC & AP & ACC & AP & ACC & AP 
& ACC & AP & ACC & AP & ACC & AP & ACC & AP 
& ACC & AP
\\
\hline

\rowcolor{gray!10} F3Net  
& 55.0 & 60.5 & 55.0 & 32.3 & 46.0 & 44.4 & 45.0 & 38.9 & 46.4 & 45.0 & 44.3 & 42.3 & 81.1 & 94.4 & 65.0 & 70.8 & 54.7 & 53.6
\\
FreqNet
& 58.4 & 70.1 & 64.1 & 78.1 & 50.0 & 33.3 & 88.3 & 98.5 & 49.8 & 56.7 & 88.9 & 99.3 & 98.7 & 99.7 & 97.8 & 99.9 & 74.5 & 79.5
\\
\rowcolor{gray!10} Fusion
& 54.3 & 59.8 & 57.1 & 35.3 & 47.2 & 45.6 & 43.8 & 37.5 & 46.9 & 50.1 & 43.6 & 41.2 & 89.5 & 97.1 & 76.3 & 79.2 & 57.3 & 51.3
\\
FatFormer
& 74.0 & 96.8 & 81.3 & 97.2 & 66.3 & 85.6 & 83.3 & 93.9 & 71.6 & 88.6 & 93.3 & 98.8 & 99.3 & 99.9 & 91.9 & 99.1 & 82.6 & 95.0
\\
\rowcolor{gray!10} AIDE
& 60.8 & 88.8 & 67.2 & 62.6 & 56.8 & 74.8 & 84.5 & 97.5 & 67.8 & 89.8 & 86.5 & 97.0 & 99.8 & 100.0 & 94.3 & 98.7 & 77.2 & 88.7
\\
VIBNet
& 70.4 & 87.3 & 95.5 & 99.5 & 68.4 & 84.1 & 65.5 & 84.2 & 77.0 & 90.2 & 95.8 & 99.4 & 98.6 & 100.0 & 93.0 & 98.5 & 83.0 & 92.9
\\
\rowcolor{gray!10} Effort
& 85.9 & 97.8 & 87.8 & 97.2 & 78.5 & 92.0 & 92.1 & 99.3 & 87.3 & 96.6 & 86.1 & 97.7 & 97.8 & 99.6 & 86.2 & 98.4 & \underline{87.7} & \underline{97.3}
\\
NPR
& 60.8 & 84.4 & 81.0 & 99.6 & 73.6 & 97.1 & 100.0 & 100.0 & 70.0 & 85.8 & 97.7 & 100.0 & 99.4 & 99.8 & 88.5 & 100.0 & 83.9 & 95.8
\\
\rowcolor{gray!10} FerretNet
& 71.9 & 93.5 & 82.5 & 96.3 & 70.0 & 87.0 & 100.0 & 100.0 & 75.8 & 92.3 & 85.3 & 98.7 & 99.3 & 100.0 & 86.9 & 98.0 & 84.0 & 95.7
\\
\rowcolor[HTML]{D7F6FF} 
\ourmethod{}
& 84.7 & 99.0 & 88.7 & 98.2 & 77.0 & 94.1 & 93.1 & 99.7 & 87.0 & 95.3 & 90.0 & 99.7 & 98.8 & 99.9 & 92.7 & 99.8 & \textbf{89.0} & \textbf{98.2}
\\

\hline
\end{tabular}}}

\end{table*}

\begin{figure*}[t]
  \centering
  \captionsetup{font=small}

  \begin{minipage}[t]{0.32\textwidth}
    \vspace{0pt}\centering
    \setlength\tabcolsep{5pt}
    \renewcommand\arraystretch{1.18}
    \begin{adjustbox}{width=\linewidth, valign=t}
      \begin{tabular}{ccccc|c}
        \hline \thickhline
        \rowcolor{gray!20}
        TCP & TSP & DRC & HAF & FGCA & Avg. \\
        \hline\hline
        \checkmark & \texttimes  & \checkmark & \checkmark & \checkmark & 78.4 \\
        \rowcolor{gray!10}
        \texttimes   & \checkmark & \checkmark & \checkmark & \checkmark & 82.4 \\
        \checkmark & \checkmark & \texttimes   &  \texttimes & \checkmark & 88.3 \\
        \rowcolor{gray!10}
        \checkmark & \checkmark & \checkmark &\texttimes   & \checkmark & 90.2 \\
        \checkmark & \checkmark & \texttimes  & \checkmark & \checkmark & 91.1 \\
        \rowcolor{gray!10}
        \checkmark & \checkmark & \checkmark & \checkmark & $\times$   & 91.2 \\
        \rowcolor[HTML]{D7F6FF}
        \checkmark & \checkmark & \checkmark & \checkmark & \checkmark & 94.4 \\
        \hline
      \end{tabular}
    \end{adjustbox}
  \end{minipage}
  \hfill
  \begin{minipage}[t]{0.67\textwidth}
    \vspace{0pt}\centering
    \includegraphics[width=\linewidth]{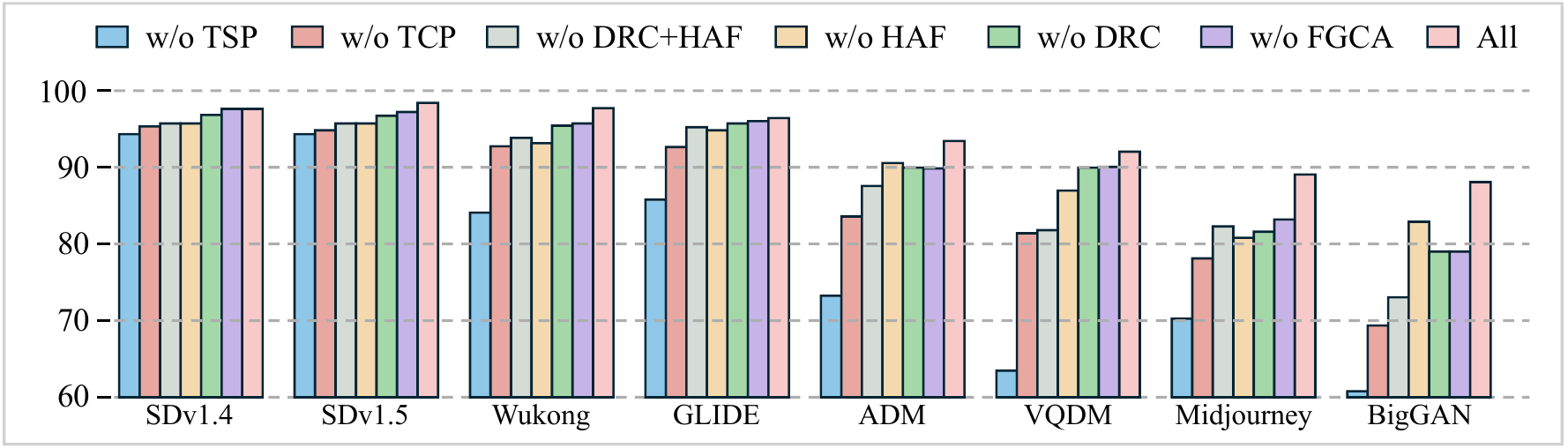}
  \end{minipage}

  \vspace{-5pt}
  \caption{Ablation study of core model components on the GenImage benchmark. 
  The left table reports average detection accuracy under different module combinations, while the right plot shows per-generator performance.}
  \vspace{-5pt}
  \label{fig:component_ablation}
\end{figure*}

\section{Experiments}
\subsection{Experimental Setup}
\noindent\textbf{Datasets.}
To evaluate the effectiveness of different methods in cross generative model and  cross dataset scenarios, we adopt the following four settings.

\begin{itemize}[leftmargin=*]

\item[\ding{182}] \textbf{{GenImage Benchmark}}~\cite{zhu2023genimage}. We train on the SDv1.4~\cite{ho2020denoising} subset and evaluate on images generated by seven diffusion models and one GAN.

\item[\ding{183}] \textbf{{DeepFaceGen Benchmark}}~\cite{bei2024large}. We adopt the official data split, partitioning the dataset into training, validation, and test sets with a ratio of $7\!:\!1\!:\!2$, this benchmark evaluates faces generated by 12 instant-guided methods.

\item[\ding{184}] \textbf{{DiffusionForensics Benchmark}}~\cite{wang2023dire}. We train on the SDv1.4 subset of GenImage and perform cross-dataset validation on DiffusionForensics across multiple diffusion generation frameworks.

\item[\ding{185}] \textbf{{COSPY Benchmark}}~\cite{cheng2025co}. We train on the SDv1.4 subset of GenImage and test on the portion of COSPY collected after 2024 to assess generalization to more recently developed generative models.
\end{itemize}

More dataset details are provided in Appendix.

\noindent\textbf{Evaluation Metrics.}
We report Accuracy (ACC), Average Precision (AP), and Area Under the ROC Curve (AUC). Across all experiments, forged images are treated as the positive class and real images as the negative class.

\noindent\textbf{Implementation Details.}
During training phase, we use the AdamW optimizer with an initial learning rate of $1\times10^{-4}$, betas = (0.9, 0.999), and weight decay of $1\times10^{-4}$. The batch size is 64. A StepLR scheduler is adopted with a decay factor of 0.7 per epoch. All experiments are implemented in PyTorch and run on a single NVIDIA RTX A6000 GPU. Additional implementation details and hyperparameters are included in the Appendix.

\subsection{Comparison with Competing Methods.}

\noindent\textbf{Performance on GenImage.}
We first evaluate cross-generator generalization on GenImage. As shown in \cref{tab:table1}, \ourmethod{} achieves the highest average ACC of 94.4\%, outperforming the strongest baseline, CKNNA, by 2.4 percentage points. Although several baselines perform strongly on individual generators, their accuracy varies substantially across models. In contrast, \ourmethod{} maintains at least 88.2\% ACC on all generators and achieves a notable advantage on ADM, indicating more balanced performance across heterogeneous generation architectures.

\noindent\textbf{Performance on DeepFaceGen.}
Following the protocol of prior work~\cite{lou2025std}, we evaluate all methods on DeepFaceGen using AUC under the official 7:1:2 training, validation, and test split. As shown in \cref{fig:radar}, \ourmethod{} achieves the best average AUC of 95.12\% and consistently strong performance across all 12 generation methods, demonstrating stable generalization across diverse face-generation paradigms.

\noindent\textbf{Performance on DiffusionForensics.}
We further evaluate cross-dataset generalization on eight diffusion generators from DiffusionForensics. As reported in \cref{tab:df_table}, \ourmethod{} achieves the best mean ACC and AP of 89.0\% and 98.2\%, respectively. Compared with the strongest baseline, Effort, it improves ACC by 1.3 and AP by 0.9 percentage points, confirming its effectiveness in capturing transferable artifacts across diffusion models.

\begin{table}
\centering
\caption{ACC performance comparison of \ourmethod{} and other methods on COSPY.}
\vspace{-10pt}
\label{tab:COSPY}
\renewcommand{\arraystretch}{1}
\resizebox{\linewidth}{!}{%
\begin{tabular}{lcccccc}
\thickhline
\rowcolor{gray!20}
Method & SegMoE & SD-3-m & PG-v2.5 & FLUX.1-sch & FLUX.1-dev & Mean \\
\midrule
\rowcolor{gray!10} NPR       & 96.2 & 67.7 & 76.6 & 82.2 & 76.9 & 79.9 \\
FerretNet & 98.7 & 76.3 & 83.1 & 94.6 & 95.1 & 89.6 \\
\rowcolor[HTML]{D7F6FF}  \ourmethod{} & 98.8 & 85.9 & 87.3 & 96.5 & 92.6 & 92.2 \\
\thickhline
\end{tabular}%
}
\vspace{-10pt}
\end{table}

\noindent\textbf{Performance on COSPY.}
As shown in \cref{tab:COSPY}, we evaluate generalization to five recent generators on the challenging COSPY benchmark. \ourmethod{} achieves the best result on four generators and the highest ACC of 92.2\%, exceeding FerretNet by 2.6 percentage points. These results demonstrate strong generalization to newly developed models.

More experimental results are provided in \textbf{Appendix}.

\begin{table*}[t]\small
\captionsetup{font=small}
\caption{Ablation study of key hyperparameters on the GenImage benchmark. We evaluate the impact of patch size, patch number, patch selection strategy, and step number on performance across eight generators. The reported values denote ACC (\%).}
\vspace{-5pt}
\centering
{\scriptsize
\resizebox{0.95\linewidth}{!}{
\setlength\tabcolsep{3pt}
\renewcommand\arraystretch{1.1}
\begin{tabular}{r|cccc|cccc|ccc|ccc}
\hline\thickhline
\rowcolor{gray!20}
& \multicolumn{4}{c|}{Patch Size} 
& \multicolumn{4}{c|}{Patch Number} 
& \multicolumn{3}{c|}{Patch Select} 
& \multicolumn{3}{c}{Step Number} \\
\cline{2-15}
\rowcolor{gray!20}
\multirow{-2}{*}{Generator}
& 8$\times$8 & 16$\times$16 & 24$\times$24 & 32$\times$32 
& 1 & 2 & 3 & 4 
& TCP+TSP & TCP+TMP & TMP+TSP 
& 2 & 3 & 4 \\
\hline

Midj 
& 78.6 & 89.2 & 87.1 & 80.2 
& 89.2 & 83.7 & 85.5 & 82.8 
& 89.2 & 80.5 & 77.9 
& 81.3 & 89.2 & 82.8 \\
\rowcolor{gray!10}
SDv1.4 
& 96.7 & 98.8 & 97.9 & 98.3 
& 98.8 & 98.3 & 98.8 & 98.5 
& 98.8 & 96.3 & 96.9 
& 99.3 & 98.8 & 98.5 \\
SDv1.5 
& 96.5 & 98.6 & 97.9 & 98.2 
& 98.6 & 98.3 & 98.6 & 98.3 
& 98.6 & 96.0 & 96.9 
& 99.2 & 98.6 & 98.3 \\
\rowcolor{gray!10}
ADM 
& 89.6 & 93.6 & 93.2 & 92.6 
& 93.6 & 93.1 & 90.4 & 94.2 
& 93.6 & 78.7 & 90.9 
& 89.7 & 93.6 & 94.2 \\
GLIDE 
& 95.4 & 96.6 & 96.8 & 94.4 
& 96.6 & 97.7 & 97.7 & 96.5 
& 96.6 & 85.9 & 93.2 
& 96.7 & 96.6 & 96.5 \\
\rowcolor{gray!10}
Wukong 
& 93.3 & 97.9 & 96.9 & 97.3 
& 97.9 & 97.0 & 97.9 & 98.0 
& 97.9 & 94.7 & 93.6 
& 98.2 & 97.9 & 98.0 \\
VQDM 
& 87.4 & 92.2 & 92.5 & 91.8 
& 92.2 & 92.7 & 90.9 & 95.2 
& 92.2 & 78.8 & 90.8 
& 91.6 & 92.2 & 95.2 \\
\rowcolor{gray!10}
BigGAN 
& 84.6 & 88.2 & 86.8 & 81.4 
& 88.2 & 86.5 & 79.0 & 84.5 
& 88.2 & 85.4 & 76.8 
& 73.1 & 88.2 & 84.5 \\
\rowcolor[HTML]{D7F6FF}
Avg. 
& 90.3 & 94.4 & 93.6 & 91.8 
& 94.4 & 93.9 & 92.4 & 93.5 
& 94.4 & 87.0 & 89.6 
& 91.1 & 94.4 & 93.5 \\
\hline
\end{tabular}}}
\label{tab:hyper_ablation}
\vspace{-5pt}
\end{table*}

\begin{figure*}[t]
\centering
\includegraphics[width=0.95\linewidth]{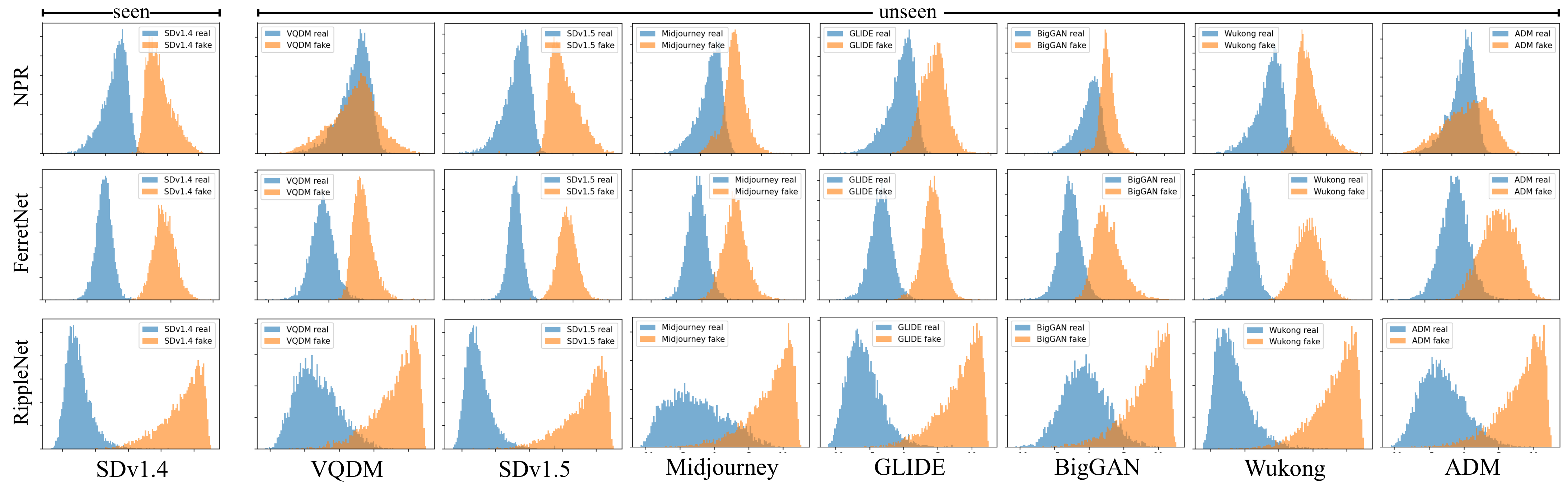}
\vspace{-10pt}
\captionsetup{font=small}
\caption{Inter-class separability comparison among \ourmethod{}, NPR and FerretNet.}
\label{fig:visualization}
\vspace{-10pt}
\end{figure*}

\subsection{Ablation Studies}

To quantify the contribution of each design, we conduct ablation studies on GenImage for Forgery-Sensitive Patch Selection~(FSPS), Structured Differential Artifact Modeling~(SDAM), and Frequency-Guided Cross-Attention~(FGCA).

\noindent\textbf{Effectiveness of FSPS.}
As shown in \cref{fig:component_ablation}, using only TCPs or TSPs reduces the average ACC from 94.4\% to 78.4\% and 82.4\%, respectively. This substantial degradation confirms that the two texture regimes provide complementary artifact evidence. Their joint use exposes differential irregularities under contrasting local statistics, providing more informative inputs for subsequent modeling.

\noindent\textbf{Effectiveness of SDAM.}
Removing DRC or HAF decreases the average ACC by 3.3 and 4.2 percentage points, respectively, while removing both yields a larger drop of 6.1 points. These results show that independent directional residuals are insufficient: DRC captures cyclic dependencies across directions, whereas HAF emphasizes informative neighborhood scales. Their combination produces more discriminative and transferable differential representations.

\noindent\textbf{Effectiveness of FGCA.}
Removing FGCA reduces the average ACC from 94.4\% to 91.2\%. This result verifies that high-frequency responses provide complementary evidence to spatial differential representations, helping the encoder capture spectral anomalies shared across generators.

\noindent\textbf{Hyperparameter Analysis.}
As reported in \cref{tab:hyper_ablation}, a patch size of $16\times16$ achieves the highest average ACC of 94.4\%, while smaller patches provide insufficient context and larger patches introduce structural interference. Selecting one TCP and one TSP performs best; increasing the patch number provides no consistent gain. Among patch combinations, TCP+TSP outperforms TCP+TMP and TMP+TSP by 7.4 and 4.8 percentage points, confirming the complementarity of the two texture extremes. Finally, $L=3$ achieves the best result, whereas fewer steps limit the receptive range and additional steps may introduce irrelevant variations.

Additional ablation studies, robustness, and computational cost analysis are provided in \textbf{Appendix}.

\subsection{Visualization}

To examine cross-generator separability, we compare the prediction-score distributions of \ourmethod{}, NPR, and FerretNet across different generators. As shown in \cref{fig:visualization}, \ourmethod{} consistently produces clearer separation between real and generated samples, with smaller distributional variations across generators. This indicates that its learned representations are less sensitive to generator-specific shifts and support more stable cross-model discrimination.

\section{Conclusion}
This paper analyzes representation bias in generated-image detection from an information-theoretic perspective and argues that effective discrimination requires suppressing dominant high-SNR semantic components while strengthening generation-mechanism deviations manifested as low-SNR forgery artifacts. Building on this insight, we propose \ourmethod{}, which uses forgery-sensitive patch selection to focus on regions where artifacts are more likely to emerge, and combines neighborhood differential modeling with fine-grained attention encoding to learn artifact representations less dependent on semantic content. Extensive experiments demonstrate that \ourmethod{} achieves stable and competitive performance across multiple cross-generator benchmarks, exhibiting strong generalization to unseen generative models.

\clearpage
\appendix

In the Supplementary Material, we first present a more systematic and formal exposition of the theoretical motivation and core design principles of \ourmethod{}. Subsequently, we provide the complete details of the experimental setup, including more comprehensive descriptions of the datasets and configurations of the training hyperparameters. Building on this foundation, we further supplement the detailed experimental results that were not included in the main text and offer an additional series of ablation studies. 

\section{Theoretical Motivation of \ourmethod{}}

Diffusion models have become the dominant paradigm for image generation, producing images with highly realistic global structure and strong low-frequency fidelity. Nevertheless, systematic discrepancies often remain in fine-scale details, especially in high-frequency components. To motivate the design of \ourmethod{}, we analyze this phenomenon from two complementary perspectives. First, we discuss why diffusion training tends to under-emphasize high-frequency components. Second, from the viewpoint of the information bottleneck, we explain why conventional detectors are easily biased toward semantic content rather than subtle forgery traces. Together, these observations motivate the architectural choices of \ourmethod{}.

\noindent{\textbf{Frequency-Domain Gradient Imbalance in Diffusion Training.}}
In diffusion models, the learning dynamics of different frequency components are closely related to their signal-to-noise ratios (SNR). Since high-frequency components typically have lower energy and are more susceptible to corruption by injected noise, they tend to contribute less effectively to optimization, which may result in persistent distortions in synthesized high-frequency details.

Consider a standard DDPM, whose objective at timestep $t$ is defined as
\begin{equation}
    \mathcal{L}(\theta)
    = \mathbb{E}_{x_0,\epsilon,t}
    \left[
        \left\| \epsilon - \epsilon_\theta(x_t,t) \right\|_2^2
    \right],
\end{equation}
where
\begin{equation}
    x_t = \sqrt{\alpha_t}\,x_0 + \sqrt{1-\alpha_t}\,\epsilon,
\end{equation}
$\alpha_t$ denotes the timestep-dependent decay coefficient, $\epsilon \sim \mathcal{N}(0,I)$, and $\epsilon_\theta$ is the noise prediction network parameterized by $\theta$.

Let $\hat{x}_0(f)$ and $\hat{\epsilon}(f)$ denote the Fourier coefficients of the clean image and the noise at frequency channel $f$, respectively. Then the noisy input at frequency $f$ can be written as
\begin{equation}
    \hat{x}_t(f)
    = \sqrt{\alpha_t}\,\hat{x}_0(f)
    + \sqrt{1-\alpha_t}\,\hat{\epsilon}(f).
\end{equation}
If $P_x(f)$ and $P_\epsilon(f)$ denote the signal and noise power at frequency $f$, the corresponding channel-wise SNR is
\begin{equation}
    \mathrm{SNR}_f
    =
    \frac{\alpha_t P_x(f)}{(1-\alpha_t) P_\epsilon(f)}.
\end{equation}

\begin{figure}[t]
    \centering
    \includegraphics[width=0.5\textwidth]{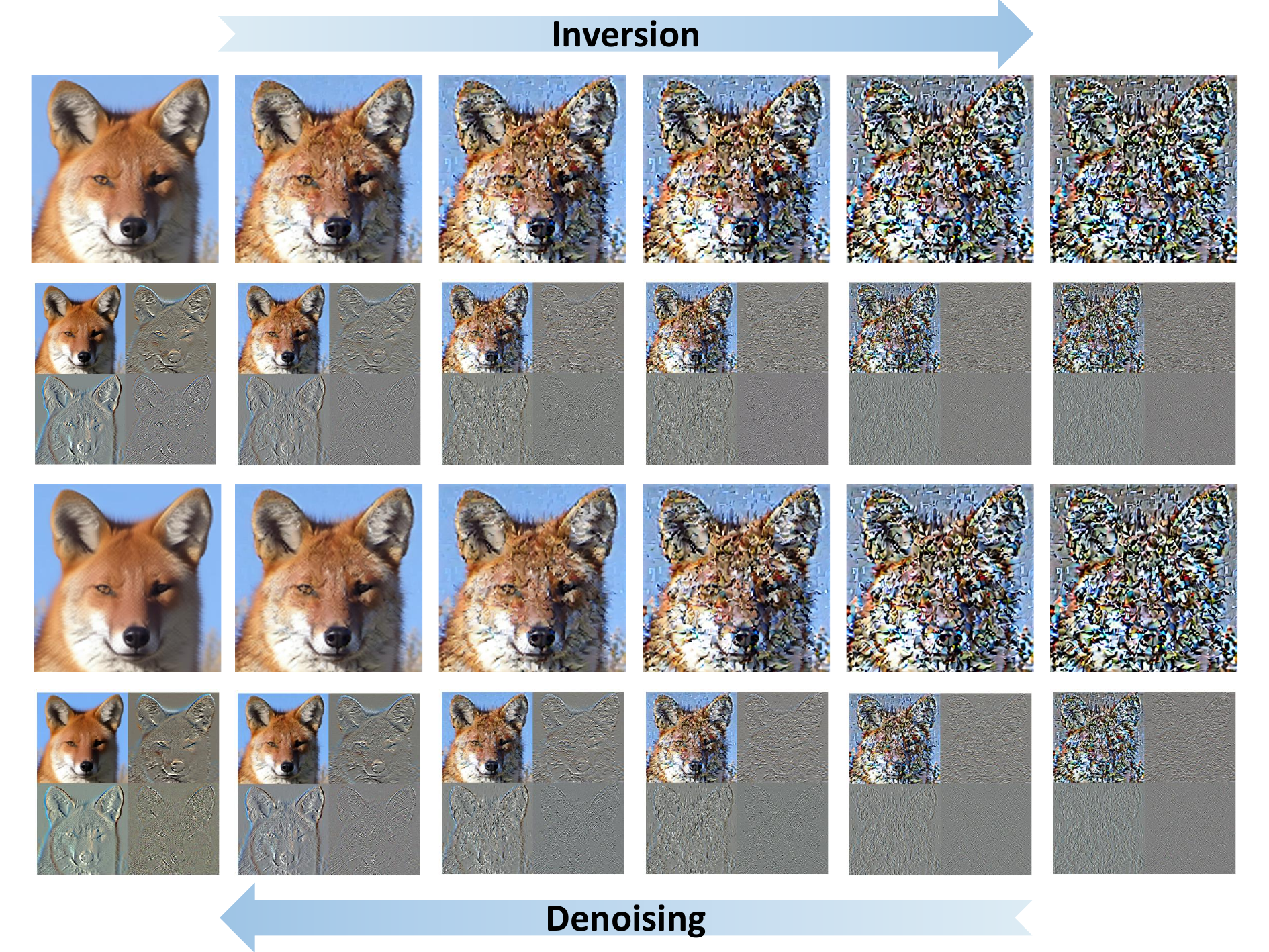}
    \caption{Wavelet decomposition exposes the high-frequency degradation occurring along diffusion reconstruction process.}
    \label{fig:Diffusion}
\end{figure}

Under the mean squared error objective, the gradient contribution of frequency channel $f$ can be approximated as
\begin{equation}
    \nabla_\theta \mathcal{L}_f
    \approx
    \mathbb{E}\Big[
        \big(\hat{\epsilon}(f)-\hat{\epsilon}_\theta(f)\big)
        \cdot
        \nabla_\theta \hat{\epsilon}_\theta(f)
    \Big].
\end{equation}
Assuming that, at the early stage of training, the scale of $\nabla_\theta \hat{\epsilon}_\theta(f)$ does not vary substantially across frequency channels, the expected gradient magnitude is mainly governed by the prediction error magnitude:
\begin{equation}
    \big\|\nabla_\theta \mathcal{L}\big\|_f
    \propto
    \sqrt{
        \mathbb{E}\left[
            \big(
                \hat{\epsilon}(f)-\hat{\epsilon}_\theta(f)
            \big)^2
        \right]
    }.
\end{equation}

Because high frequency components generally exhibit lower signal energy and are more easily submerged in noise during the forward diffusion process, their effective SNR is typically lower. This suggests that
\begin{equation}
    \big\|\nabla_\theta \mathcal{L}\big\|_f
    \propto
    \mathrm{SNR}_f^{1/2},
\end{equation}
implying that high frequency channels tend to provide weaker optimization signals during training. As a consequence, the reverse process may preserve low-frequency structure well while leaving more detectable inconsistencies in high-frequency details.

\noindent{\textbf{Detector Bias from an Information Bottleneck Perspective.}}
Beyond the intrinsic spectral properties of generative models, the representation bias of the detector itself also affects how forgery traces are exploited. Let $X$ denote the input image, $Y$ the label, and $Z=h_\theta(X)$ the intermediate representation extracted by a detector. From the information bottleneck perspective, representation learning can be approximately characterized as
\begin{equation}
    \max_\theta \; I(Z;Y) - \beta I(Z;X),
\end{equation}
where $I(\cdot;\cdot)$ denotes mutual information and $\beta$ controls the degree of compression.

From a structural perspective, the image can be decomposed into a semantic component $X_s$ and a residual component $X_r$:
\begin{equation}
    X = X_s + X_r, \qquad \langle X_s, X_r \rangle = 0.
\end{equation}
Here, $X_s$ mainly captures low-frequency structure and semantic layout, while $X_r$ contains fine-grained residual variations, including subtle textural and statistical irregularities. In standard discriminative learning, $X_s$ usually exhibits stronger and more stable correlation with the label, and therefore tends to dominate the learned representation:
\begin{equation}
    I(Z;X_s) \gg I(Z;X_r).
\end{equation}
Correspondingly, the detector is often much more sensitive to perturbations along semantic directions than residual ones:
\begin{equation}
    \big\|\nabla_{X_s} h_\theta(X)\big\|
    \gg
    \big\|\nabla_{X_r} h_\theta(X)\big\|.
\end{equation}

This bias is problematic for generated image detection. Although semantic structure is often visually salient, the cues that generalize across generators are more likely to reside in weak residual signals rather than dominant content patterns. As a result, detectors that rely primarily on semantic features may understand image content well but still fail to consistently capture generator-specific artifacts, leading to limited cross domain generalization.

\noindent{\textbf{Design Motivation of \ourmethod{}.}}
The above analysis directly motivates the design of \ourmethod{}, which aims to suppress semantic dominance and enhance the modeling of weak forgery related residual cues. Specifically, \textbf{FSPS} selects forgery sensitive regions to reduce semantically redundant content at the input stage. \textbf{SDAM} explicitly models directional statistical dependencies within local pixel neighborhoods, transforming subtle residual cues into learnable structural representations. \textbf{FAE} further performs pixel-level attention over these residual patterns and injects frequency guided high frequency priors into the early stages of pixel-level MHSA. Through this design, \ourmethod{} encourages the detector to focus on fine-grained statistical inconsistencies that are weakly tied to image semantics but more closely related to the underlying generative process, thereby improving generalization across generators and datasets.

\begin{table*}[t]\small
\centering
\captionsetup{font=small}
\caption{AUC (\%) comparison of \ourmethod{} and other forgery detection models on the DeepFaceGen. }
\label{tab:gen_table}
{\scriptsize
\resizebox{\linewidth}{!}{
\setlength\tabcolsep{3pt}
\renewcommand\arraystretch{1.15}
\begin{tabular}{r|ccccccccccccc}
\hline\thickhline
\rowcolor{gray!20}
 & Midjourney & DALL-E1 & DALL-E3 & Wenxin & SD1 & SDXLR & OJ & pix2pix & SD2 & SDXL & VD & DF-GAN & Average \\
\hline

\rowcolor{gray!10} Xception
& 77.01 & 75.45 & 86.59 & 86.87 & 87.64 & 84.13 & 89.72 & 83.42 & 87.79 & 86.06 & 85.02 & 95.42 & 85.42
\\
EfficientNet
& 79.52 & 81.74 & 88.41 & 84.34 & 86.83 & 93.46 & 89.00 & 77.61 & 85.91 & 87.65 & 83.84 & 96.71 & 86.25
\\
\rowcolor{gray!10} F3Net
& 81.65 & 84.73 & 87.23 & 89.28 & 90.95 & 86.40 & 92.72 & 81.21 & 89.11 & 91.43 & 89.52 & 93.45 & 88.14
\\
RECCE
& 87.64 & 83.25 & 89.17 & 92.13 & 89.83 & 90.46 & 96.90 & 89.71 & 95.43 & 96.75 & 95.67 & 93.54 & 91.70
\\
\rowcolor{gray!10} DNADet
& 93.44 & 85.62 & 83.90 & 91.60 & 92.40 & 89.03 & 94.04 & 88.52 & 92.70 & 92.28 & 89.21 & 94.22 & 90.58
\\
FreqNet
& 85.69 & 84.25 & 87.13 & 91.98 & 90.94 & 89.40 & 93.08 & 88.66 & 90.92 & 92.61 & 96.55 & 97.11 & 90.69
\\
\rowcolor{gray!10} DIRE
& 87.01 & 86.65 & 87.84 & 92.84 & 92.35 & 86.61 & 91.28 & 89.01 & 89.54 & 91.54 & 98.78 & 90.32 & 90.69
\\
DRCT
& 89.78 & 89.91 & 88.05 & 92.72 & 93.56 & 89.51 & 92.45 & 91.51 & 90.41 & 91.01 & 94.25 & 92.54 & 90.48
\\
\rowcolor{gray!10} UnivFD
& 88.67 & 87.64 & 89.21 & 90.01 & 90.01 & 89.01 & 88.01 & 89.54 & 91.45 & 91.01 & 89.68 & 95.98 & 91.30
\\
NPR
& 89.01 & 89.54 & 89.41 & 92.35 & 90.12 & 88.64 & 90.28 & 89.30 & 90.01 & 89.87 & 91.01 & 98.88 & 89.85
\\
\rowcolor{gray!10} STD-FD
& 94.36 & 91.21 & 90.01 & 92.90 & 93.77 & 93.50 & 97.05 & 92.34 & 96.71 & 97.05 & 100.00 & 100.00 & 94.90
\\
\rowcolor[HTML]{D7F6FF} 
RippleNet
& 94.10 & 92.95 & 92.20 & 93.10 & 96.05 & 94.90 & 96.60 & 92.00 & 96.90 & 96.80 & 96.00 & 99.80 & 95.12
\\

\hline
\end{tabular}}}
\end{table*}

\section{Experimental Setup Details}

\subsection{More Dataset Details}

\noindent{\textbf{GenImage}}~\cite{zhu2023genimage}.
GenImage contains forged images generated by eight generative models, along with an equal number of real images randomly sampled from ImageNet~\cite{deng2009imagenet}. Among these generators, seven are diffusion-based, including SDv1.4~\cite{rombach2022high}, SDv1.5~\cite{rombach2022high}, Midjourney~\cite{midjourney2022}, ADM~\cite{dhariwal2021diffusion}, GLIDE~\cite{nichol2021glide}, VQDM~\cite{gu2022vector}, and Wukong~\cite{wukong2022}, while the remaining one, BigGAN~\cite{brock2018large}, is GAN-based. Following the standard benchmark protocol, we use 162{,}000 forged images generated by SDv1.4 and an equal number of real images for training, and evaluate all methods on the official test subsets to ensure a fair assessment of cross-model generalization.

\noindent{\textbf{DeepFaceGen}}~\cite{bei2024large}.
DeepFaceGen is a large-scale face forgery benchmark. Following prior evaluation protocols, we select forged face samples generated by recent on-the-fly guidance methods, covering 12 generative frameworks, including DALL$\cdot$E~\cite{ramesh2021zero} and Stable Diffusion (SD)~\cite{rombach2022high}. The training, validation, and test splits strictly follow the official benchmark configuration to ensure fair and comparable evaluation.

\noindent{\textbf{DiffusionForensics}}~\cite{wang2023dire}.
DiffusionForensics is a benchmark designed to evaluate the generalization ability of detectors across diverse diffusion-based image generators. It covers multiple representative diffusion frameworks, including ADM, DDPM, IDDPM~\cite{nichol2021improved}, LDM, PNDM~\cite{liu2022pseudo}, VQ-Diffusion, SDv1, and SDv2, while real images are collected from LSUN and ImageNet. Owing to the diversity of diffusion architectures and sampling paradigms involved, this benchmark provides a suitable testbed for assessing whether a detector can capture generator-agnostic forensic cues rather than overfitting to artifacts specific to a single model family.

\noindent{\textbf{COSPY}}~\cite{cheng2025co}. COSPY is a challenging benchmark for evaluating the generalization ability of generated-image detectors. In our experiments, rather than using all generators in COSPY, we specifically select those released after 2024 to assess model generalizability under more recent generative advances. The selected generators include SegMoE-SD~\cite{gupta2024progressive}, SD-3-medium~\cite{rombach2022high}, PG-v2.5-1024~\cite{li2024playground}, FLUX.1-schnell~\cite{flux}, and FLUX.1-dev.

\subsection{More Hyperparameter Details}

\ourmethod{} consists of three core modules: FSPS (Forgery-Sensitive Patch Selection), SDAM (Structured Differential Artifact Modeling), and FAE (Forgery-Aware Encoder). In the FSPS stage, we first convert the input image to grayscale and partition it into non overlapping $16 \times 16$ patches. We then select the patch with the highest texture complexity and the patch with the lowest texture complexity from the sorted texture complexity list as forgery sensitive patches, forming the sets $\mathcal{P}_{\text{TC}}$ and $\mathcal{P}_{\text{TS}}$, respectively. Discrete Wavelet Transform (DWT) is applied to these regions, and the HH high-frequency subband is extracted to capture potential forgery traces.
In the SDAM module, for each pixel within each forgery-sensitive patch, we construct multi-directional and multi-scale residual features using eight directions and step size $L = 3$, explicitly modeling the local inconsistencies introduced by the generative process.
In the FAE stage, we apply 2D Rotary Position Embedding (RoPE) to pixel-level tokens and stack one layer of Frequency Guided Cross Attention (FGCA) and four layers of pixel-level Multi-Head Self-Attention (MHSA) to jointly exploit forgery cues from both the pixel and frequency domains.

\begin{table*}[t]\small
\centering
\captionsetup{font=small}
\caption{ACC (\%) and AP (\%) comparison of \ourmethod{} and other forgery detection models on the  Ojha dataset.
All methods are trained on GenImage/SDv1.4.
\textbf{Bold} indicates the best result, and \underline{underline} denotes the second-best.}
\label{tab:ojha_table}
{\scriptsize
\resizebox{\linewidth}{!}{
\setlength\tabcolsep{3pt}
\renewcommand\arraystretch{1.15}
\begin{tabular}{r|cc|cc|cc|cc|cc|cc|cc|cc|cc}
\hline\thickhline
\rowcolor{gray!20}
 & \multicolumn{2}{c|}{DALLE}
 & \multicolumn{2}{c|}{Glide\_100\_10}
 & \multicolumn{2}{c|}{Glide\_100\_27}
 & \multicolumn{2}{c|}{Glide\_50\_27}
 & \multicolumn{2}{c|}{Guided}
 & \multicolumn{2}{c|}{LDM\_100}
 & \multicolumn{2}{c|}{LDM\_200}
 & \multicolumn{2}{c|}{LDM\_200\_cfg}
 & \multicolumn{2}{c}{Mean}
\\
\cline{2-19}
\rowcolor{gray!20}
Methods
& ACC & AP & ACC & AP & ACC & AP & ACC & AP 
& ACC & AP & ACC & AP & ACC & AP & ACC & AP 
& ACC & AP
\\
\hline

\rowcolor{gray!10} F3Net  
& 69.0 & 79.8 & 47.5 & 45.1 & 47.3 & 43.8 & 47.0 & 41.0 & 51.9 & 54.4 & 65.9 & 77.6 & 63.5 & 77.0 & 66.1 & 78.7 & 57.3 & 62.2
\\
FreqNet
& 52.7 & 81.7 & 71.5 & 85.0 & 71.9 & 94.7 & 74.6 & 95.3 & 57.0 & 64.9 & 94.8 & 99.4 & 94.2 & 99.4 & 92.2 & 99.1 & 76.1 & 89.9
\\
\rowcolor{gray!10} FatFormer
& 83.0 & 95.7 & 85.4 & 96.5 & 84.6 & 96.4 & 84.8 & 96.4 & 65.5 & 88.9 & 92.6 & 98.9 & 92.4 & 99.0 & 89.8 & 97.7 & 84.8 & 96.2
\\
AIDE
& 78.7 & 95.5 & 93.0 & 99.1 & 92.4 & 99.0 & 93.0 & 99.0 & 60.7 & 93.4 & 97.0 & 99.7 & 96.2 & 99.6 & 97.0 & 98.2 & 88.5 & 97.9
\\
Fusion
& 59.1 & 79.4 & 63.0 & 77.2 & 48.7 & 44.8 & 87.5 & 97.9 & 53.5 & 67.3 & 89.4 & 98.7 & 90.9 & 97.5 & 91.2 & 97.8 & 72.9 & 82.6
\\
\rowcolor{gray!10} VIB-Net
& 93.2 & 98.0 & 72.1 & 90.6 & 71.0 & 90.3 & 68.8 & 89.4 & 73.2 & 92.5 & 94.5 & 98.3 & 94.7 & 98.5 & 87.0 & 95.7 & 81.8 & 94.2
\\
Effort
& 90.0 & 95.4 & 91.1 & 96.9 & 91.0 & 96.7 & 90.8 & 96.7 & 83.7 & 98.0 & 90.0 & 96.4 & 89.7 & 95.7 & 90.0 & 96.5 & 89.5 & 96.5
\\
\rowcolor{gray!10} NPR
& 77.2 & 93.7 & 91.2 & 99.9 & 90.4 & 99.8 & 92.5 & 99.8 & 62.5 & 80.9 & 86.2 & 98.8 & 86.4 & 98.9 & 85.5 & 99.1 & 84.0 & 96.4
\\
FerretNet
& 88.5 & 96.8 & 96.0 & 99.3 & 95.5 & 99.1 & 95.5 & 99.2 & 75.4 & 97.0 & 97.2 & 99.0 & 98.2 & 100.0 & 98.2 & 99.9 & \underline{93.1} & \textbf{98.8}
\\
\rowcolor[HTML]{D7F6FF} 
\ourmethod{}
& 95.6 & 98.5 & 95.0 & 97.8 & 95.2 & 98.6 & 94.5 & 98.6 & 83.9 & 99.0 & 95.6 & 98.8 & 95.5 & 99.0 & 94.4 & 98.2 & \textbf{93.7} & \underline{98.6}
\\

\hline
\end{tabular}}}
\end{table*}

\begin{figure*}[t]
\centering
\includegraphics[width=\linewidth]{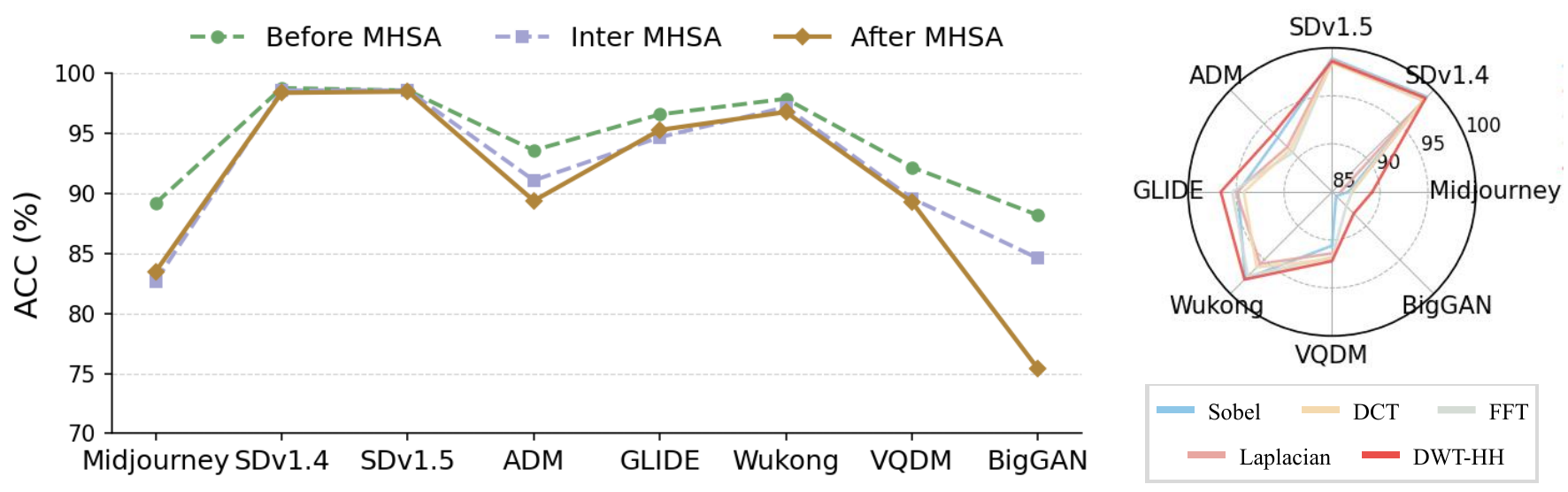}
\captionsetup{font=small}
\caption{Ablation studies on \ourmethod{}. Left: comparison of FGCA placement at different stages of the pixel-level MHSA (Before MHSA, Inter MHSA, After MHSA) across multiple GenImage test subsets. Right: comparison of different high-frequency extraction methods, including Sobel, Laplacian, DCT, FFT, and DWT-HH.}
\vspace{-10pt}
\label{fig:ablusion2}
\end{figure*}

\section{Additional Experimental Results}

\subsection{Detailed Results on DeepFaceGen}

In the main text, we presented a radar chart illustrating the overall performance of \ourmethod{} and various competitive detection methods on the DeepFaceGen benchmark, providing an intuitive comparison of their AUC metrics. In this section, we further provide the complete numerical AUC results for more fine-grained quantitative analysis. The AUC scores of all methods across the DeepFaceGen subsets are summarized in \cref{tab:gen_table}.

\subsection{Experimental Results on Ojha}

Using the SDv1.4 subset of GenImage for training, we further evaluate the cross model generalization of different detectors on the test sets introduced by Ojha et al.~\cite{ojha2023towards}. This benchmark contains forged images generated by ADM, GLIDE, DALL$\cdot$E, and LDM under different parameter settings and sampling strategies, together with real images randomly sampled from LAION~\cite{schuhmann2022laion} and ImageNet. As shown in \cref{tab:ojha_table}, we report the detailed performance of \ourmethod{} and several advanced baselines on this cross model benchmark. The results show that \ourmethod{} consistently maintains strong performance across diverse diffusion models and sampling settings, substantially outperforming existing detectors and demonstrating superior generalization ability.

\subsection{More Ablation Studies}

\noindent{\textbf{FGCA Placement.}}
To assess the optimal insertion point of the FGCA within the pixel-level encoder, we fix the number of FGCA layers and place it at different positions: before, inter, or after the pixel-level MHSA stacks. Results on GenImage are shown in \cref{fig:ablusion2}. When FGCA is inserted before the first MHSA layer (Before-MHSA), the model achieves the highest ACC on GenImage. In contrast, placing FGCA in the middle or the final stage leads to noticeable performance degradation. We attribute this to the fact that introducing frequency-domain information at an early stage enables effective frequency renormalization of pixel-level tokens, providing more forgery-aware representations for the subsequent MHSA layers. Conversely, late-stage correction cannot fully rectify the representations already formed in earlier layers, yielding limited benefits overall.

\noindent{\textbf{High-frequency Feature Extraction.}}
To identify the most suitable high frequency extraction operator for FGCA, we compare several commonly used choices, including Sobel, Laplacian, FFT, DCT, and the DWT adopted in our model. As shown in \cref{fig:ablusion2}, spatial operators such as Sobel and Laplacian can capture local edge variations, but are limited in characterizing the broader frequency-domain artifacts introduced by generative models. Although FFT and DCT provide strong frequency representations, they lack explicit spatial localization, making it difficult to align the extracted high-frequency cues with pixel-level tokens. In contrast, DWT offers a more balanced spatial frequency decomposition. In particular, its HH subband can more effectively capture transferable high frequency forgery traces while preserving spatial correspondence with pixel-level features, thereby achieving the best performance among all operators.

\subsection{Robustness to Common Post-processing}

Common post-processing operations can modify or suppress forensic traces, making robustness a persistent challenge for generated-image detection. Importantly, this sensitivity is not specific to \ourmethod{}, but represents a common limitation among lightweight, artifact-oriented detectors, whose discriminative evidence can be readily altered by compression, resampling, and smoothing. Recent large-scale evaluations have similarly revealed substantial limitations of existing detectors under real-world image transformations~\cite{li2025bridging}. Achieving stronger robustness therefore often requires dedicated degradation-aware training or representation design.

To examine this issue, we evaluate resizing with scale factors of $0.75$ and $1.25$, rotation by $45^{\circ}$, JPEG compression with quality factors of $95$ and $75$, and Gaussian blur with kernel sizes of $3$ and $5$. None of the compared methods uses post-processing augmentation during training. The experiment therefore measures their inherent zero-shot sensitivity to unseen transformations rather than robustness obtained through dedicated optimization.

\begin{table}[t]
\centering
\captionsetup{font=small}
\caption{Evaluation under common post-processing on GenImage in terms of AP~(\%). None of the methods uses post-processing augmentation during training. \textbf{Bold} and \underline{underline} denote the best and second-best results, respectively.}
\label{tab:postprocessing}
\renewcommand{\arraystretch}{1.05}
\setlength{\tabcolsep}{3.5pt}
\resizebox{\linewidth}{!}{%
\begin{tabular}{lcccccccc}
\hline\thickhline
\multirow{2}{*}{Method}
& \multirow{2}{*}{Clean}
& \multicolumn{2}{c}{Resize}
& \multicolumn{1}{c}{Rotation}
& \multicolumn{2}{c}{JPEG}
& \multicolumn{2}{c}{Gaussian Blur} \\
\cmidrule(lr){3-4}
\cmidrule(lr){5-5}
\cmidrule(lr){6-7}
\cmidrule(lr){8-9}
& & $S{=}0.75$ & $S{=}1.25$
& $45^{\circ}$
& $Q{=}95$ & $Q{=}75$
& $K{=}3$ & $K{=}5$ \\
\midrule
NPR
& 95.8 & 83.6 & 81.2 & 90.2
& 56.1 & 48.7 & 68.4 & 62.9 \\

FerretNet
& \underline{99.0} & 93.8 & 94.4 & \underline{97.9}
& 60.8 & 49.6 & 73.6 & 71.0 \\

Effort
& 97.9 & \underline{94.6} & \underline{95.3} & 96.7
& \textbf{82.4} & \textbf{75.6} & \textbf{86.6} & \textbf{82.2} \\

\ourmethod{}
& \textbf{99.7} & \textbf{96.6} & \textbf{95.8} & \textbf{98.7}
& \underline{67.5} & \underline{54.8} & \underline{79.9} & \underline{74.3} \\
\hline\thickhline
\end{tabular}%
}
\end{table}

As shown in \cref{tab:postprocessing}, all evaluated lightweight detectors exhibit varying degrees of degradation after post-processing, with JPEG compression and Gaussian blur causing the most pronounced changes. These operations directly suppress or reorganize the high-frequency and local statistical traces used by artifact-oriented detectors. The results therefore reflect a broader limitation of this class of efficient detection methods rather than a weakness unique to the proposed framework. Dedicated transformation-aware augmentation or consistency regularization could further improve robustness and remains complementary to our representation design.

\begin{table}[t]
\centering
\caption{\textbf{Computational cost overhead.}}
\label{tab:efficiency}
\renewcommand{\arraystretch}{1}
\resizebox{\linewidth}{!}{%
\begin{tabular}{lcccc}
\hline
 & FerretNet & DIRE & UnivFD & \ourmethod{} \\
\midrule
Parameters (M)        & 1.4  & 23.5 & 85.5 & 8.6 \\
GFLOPs                & 1.74 & 4.1  & 17.6 & 4.28 \\
Inference Time (ms)   & 22.9 & 2425.4 & 93.8 & 46.5 \\
GenImage (ACC,\%)    & 89.7 & 71.2 & 91.1 & 94.4 \\
\hline
\end{tabular}%
}
\end{table}

\section{Computational Overhead.}
\Cref{tab:efficiency} compares \ourmethod{} with representative baselines in terms of model size, FLOPs, inference latency, and detection accuracy. \ourmethod{} achieves the highest GenImage accuracy (94.4\%) with only 8.6M parameters, remaining far smaller than large-scale pretrained detectors such as UnivFD (85.5M). Its computational cost is also moderate (4.28 GFLOPs), substantially lower than UnivFD (17.6 GFLOPs) and far more efficient than reconstruction based methods such as DIRE, whose inference is considerably slower. While \ourmethod{} is moderately heavier than FerretNet, it improves accuracy by 4.7 percentage points, showing that the gain comes from more effective forgery modeling rather than simply scaling up computation. In general, \ourmethod{} offers a strong trade-off of accuracy and efficiency, suggesting favorable practical efficiency.

\end{document}